\documentclass[conference]{IEEEtran}
\IEEEoverridecommandlockouts
\usepackage{cite}
\usepackage{amsmath,amssymb,amsfonts}
\usepackage{algorithmic}
\usepackage{graphicx}   
\usepackage{textcomp}
\usepackage{xcolor}
\def\BibTeX{{\rm B\kern-.05em{\sc i\kern-.025em b}\kern-.08em
    T\kern-.1667em\lower.7ex\hbox{E}\kern-.125emX}}

\usepackage{multirow}
\usepackage{booktabs}
\usepackage{array}
\usepackage{color}
\usepackage{colortbl}
\usepackage{rotating}
\usepackage{bbding}
\usepackage{pifont}
\newcommand{\tabincell}[2]{\begin{tabular}{@{}#1@{}}#2\end{tabular}}
\begin{document}


\title{GUIDED FOCAL STACK REFINEMENT NETWORK FOR LIGHT FIELD \\ SALIENT OBJECT DETECTION

\thanks{$^*$Corresponding author: Keren Fu.}
}

\makeatletter
\newcommand{\linebreakand}{%
  \end{@IEEEauthorhalign}
  \hfill\mbox{}\par
  \mbox{}\hfill\begin{@IEEEauthorhalign}
}
\makeatother


\author{\IEEEauthorblockN{Bo Yuan, 
Yao Jiang, 
Keren Fu$^*$, 
Qijun Zhao}
\IEEEauthorblockA{\textit{College of Computer Science, Sichuan University}\\
Chengdu 610065, China}}

\maketitle

\begin{abstract}
Light field salient object detection (SOD) is an emerging research direction attributed to the richness of light field data. However, most existing methods lack effective handling of focal stacks, therefore making the latter involved in a lot of interfering information and degrade the performance of SOD. To address this limitation, we propose to utilize multi-modal features to refine focal stacks in a guided manner, resulting in a novel guided focal stack refinement network called GFRNet. 
To this end, we propose a guided refinement and fusion module (GRFM) to refine focal stacks and aggregate multi-modal features. In GRFM, all-in-focus (AiF) and depth modalities are utilized to refine focal stacks separately, leading to two novel sub-modules for different modalities, namely AiF-based refinement module (ARM) and depth-based refinement module (DRM). 
Such refinement modules enhance structural and positional information of salient objects in focal stacks, and are able to improve SOD accuracy. 
Experimental results on four benchmark datasets demonstrate the superiority of our GFRNet model against 12 state-of-the-art models.
\end{abstract}


\section{Introduction}\label{sec:intro}
Salient object detection (SOD) \cite{1,2} is an essential and important task, which aims to locate and segment the most attractive regions in a scene. It has been applied to various computer vision tasks, such as object detection and recognition\cite{3}, semantic segmentation\cite{5} and image captioning \cite{7}. According to different input data types, SOD can be categorized into 2D (RGB) \cite{rgb2, rgb9, rgb10}, 3D (RGB-D) \cite{rd6,rd7} and 4D (light field) \cite{fs1,fs9,fs11} SOD. Recently, 4D (light field) SOD has attracted increasing interest, as it can extract saliency cues from light field data containing rich information.  

Light field SOD datasets \cite{fs2,fs3,fs4,fs6} typically convert raw light field data into several formats to use, mainly including all-in-focus (AiF) images, focal stacks, and depth maps. Although focal stacks, which consist of slices of various focused depth, contain rich light field cues, interfering information in focal stacks may degrade the performance of SOD. Such interference often comes from those focal slices that focus on irrelevant depth levels, where in-focus regions are usually background while blurred regions in turn contain desired objects, easily causing ``false politive'' and ``false negetive'' in the final detection. 
Unfortunately, most existing methods lack effective handling of focal stacks. Fig.\:\ref{fig_strategy} summarizes three typical ways of focal stack handling. In Fig.\:\ref{fig_strategy}(a), representative methods (\emph{e.g.} \cite{fs4,fs12,fs14}) directly fuse focal slices into a single feature before merging it with other modalities in the subsequent decoder. In Fig.\:\ref{fig_strategy}(b), related methods \cite{fs7,fs8} take AiF/depth and focal slice features as a whole, and aggregate them directly. 
Although specific techniques, \emph{e.g.} ConvLSTM \cite{fs4, fs7, fs8}, 2D convolution \cite{fs14}, and 3D convolution \cite{fs12} are utilized for focal slice fusion/aggregation, interfering information is still hardly suppressed via conventional convolutions. This issue is even worse when focal stack features have already been merged/fused into a single one. To address this, we adopt the idea in Fig.\:\ref{fig_strategy}(c), which uses the other modalities (\emph{e.g}., all-in-focus (AiF), depth) to guide and refine the focal stack before fusing slices into a single feature. Through this way, interference lying in the focal stack can be better purified.
It should be noted that a recent work \cite{fs11} also adopts this strategy, which uses AiF information to enhance focal stack features by graph neural networks. Different from \cite{fs11}, we propose to refine the focal stack under the guidance of multi-modal features (both AiF and depth), and design novel refinement modules according to modality-specific properties.

\begin{figure}
\centerline{\includegraphics[width=1.0\columnwidth]{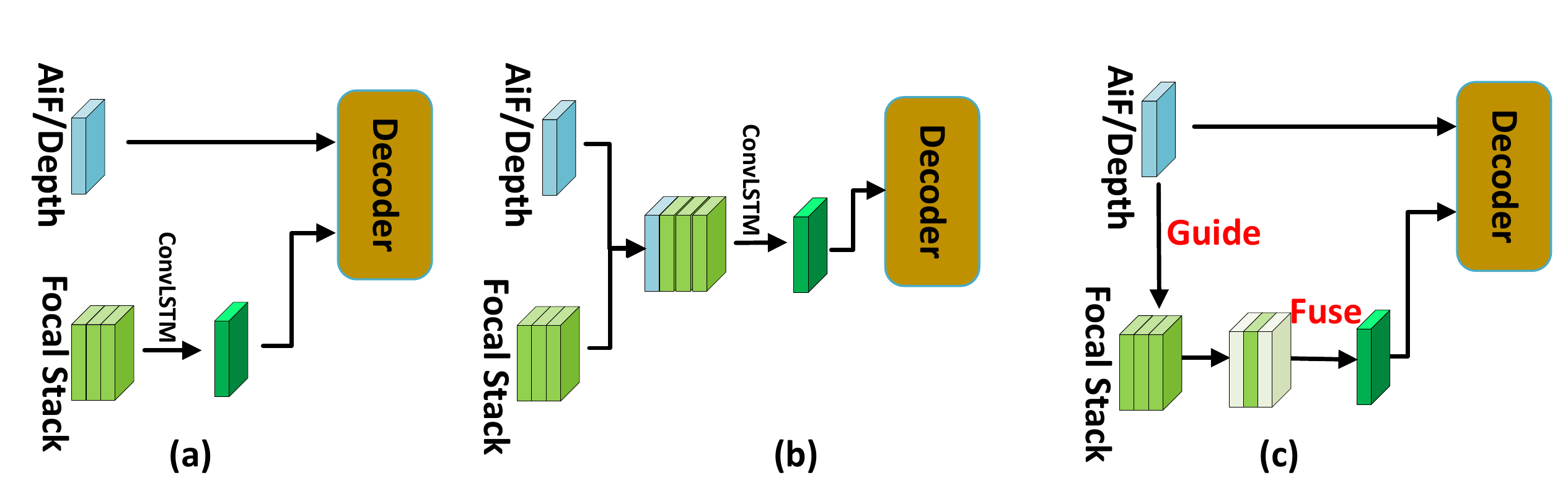}}
\caption{\small Three typical ways of focal stack handling, and the proposed model belongs to (c).}
\label{fig_strategy}
\vspace{-0.6cm}
\end{figure}

In this paper, we propose a novel framework named guided focal stack refinement network (GFRNet), which can effectively refine focal stacks in a guided manner and improve light field SOD. Specifically, we integrate a guided refinement and fusion module (GRFM), which includes two functions, \emph{i.e.}, focal stack refinement and multi-modal aggregation. Further in GRFM, we propose two sub-modules for guided refinement, namely AiF-based refinement module (ARM) and depth-based refinement module (DRM). Because of different characteristics of AiF images and depth maps, we adopt two distinct strategies for the two sub-modules, with enhancement in terms of spatial structure details and positional information, respectively.
In all, our contributions in this paper are three-fold:

\begin{itemize}
\setlength{\itemsep}{0pt}
    \setlength{\parsep}{0pt}
    \setlength{\parskip}{0pt}
    \setlength{\topsep}{0pt}
    \item We propose GFRNet based on the pioneering idea of leveraging AiF and depth cues to guide the refinement of focal stacks, which is conducive to light field SOD.
    
    \item We propose a new guided refinement and fusion module (GRFM) to refine focal stacks and aggregate different modalities. According to modality-specific properties, we design AiF-guided and depth-guided refinement strategies and sub-modules, namely AiF-based refinement module (ARM) and depth-based refinement module (DRM).
    
    \item Extensive experiments show the superiority of our GFRNet against 12 state-of-the-art light field SOD methods, and the necessity of employing modality-specific refinement modules is also validated.

\end{itemize}

\begin{figure*}
\centerline{\includegraphics[width=\textwidth]{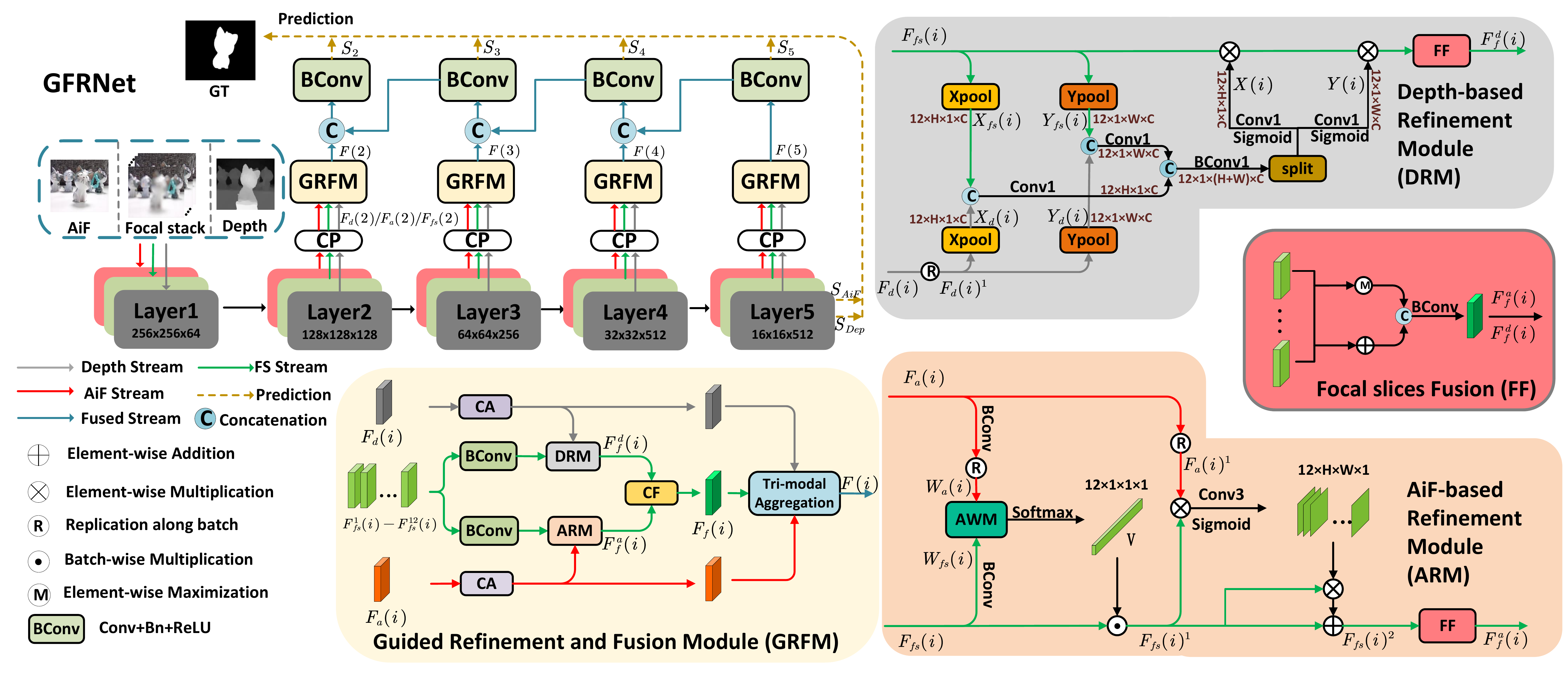}}\vspace{-0.3cm}
\caption{\small Overview of GFRNet. The overall architecture is shown in the upper left. ``CP'' is the compression module, ``CA'' denotes the channel attention module, ``CF'' denotes the cross-fusion operation, and ``AWM'' indicates the alignment weighting module.  Details of these modules/operations are described in Sec.\:\ref{sec:method}.}
\label{fig_blockdiagram}
\vspace{-0.3cm}
\end{figure*}

\section{Related Work}
Traditional light field SOD methods have proven the effectiveness of using light field data. Li \emph{et al.} \cite{fs_tra1} proposed the earliest light field SOD method, and it conducted the first benchmark dataset. After that, Li \emph{et al.} \cite{fs_tra2} constructed a saliency dictionary based on weighted sparse coding to achieve saliency maps. In addition, several methods utilized relative locations \cite{fs_tra4}, background priors \cite{fs_tra6} and dark channel priors \cite{fs_tra5} for this task. 

With the development of deep learning technology, deep neural networks have been used to boost the performance of SOD. A few deep learning-based light field SOD methods have improved the quality of saliency maps. Although raw light field data can be converted into several data formats, the majority of deep light field SOD methods have adopted focal stacks combined with other data formats (\emph{e.g.,} AiF images, depth maps) to achieve detection. Both Wang \emph{et al.} \cite{fs4} and Wang \cite{fs14} employed separate networks to process features of different modalities, and fused saliency predictions of different networks to get final results. However, the former utilized two networks to process AiF and focal stack features, respectively, while the latter added depth maps as the third modal data. Zhang \emph{et al.} \cite{fs12} utilized 3D convolution to fuse focal slices before interaction with AiF features. Piao \emph{et al.} \cite{fs10} explored focal slices in a region-wise way and integrated focused salient regions. Liang \emph{et al.} \cite{fs15} proposed a weakly-supervised learning framework for light field SOD. Moreover, Zhang \emph{et al.} \cite{fs7}, Piao \emph{et al.} \cite{fs9} and Zhang \emph{et al.} \cite{fs8} all fused different slices in the focal stack by ConvLSTM. However, Zhang \emph{et al.} \cite{fs7} and Zhang \emph{et al.} \cite{fs8} first cascaded AiF and focal slice features along batch, and employed varying attention weights for different slices, while Piao \emph{et al.} \cite{fs9} used ConvLSTM to fuse only focal slices and adopted knowledge distillation to improve feature extraction of the AiF branch. In addition to the above methods, Liu \emph{et al.} \cite{fs11} proposed to adopt graph neural networks to model the relationship between focal slices and AiF images, which can improve the focal slice fusion process.

Different from the above methods, our GFRNet refines the focal stack under the guidance of multi-modal features (AiF combined with depth features) to boost SOD, and utilizes specific refinement strategies for different modalities.

\section{Proposed Method}\label{sec:method}
The overview of our proposed GFRNet is illustrated in Fig.\:\ref{fig_blockdiagram}, which is a three-stream UNet-like architecture. Following previous works \cite{fs7,fs11}, each encoder stream adopts the VGG-19 \cite{16} network. We only use the side outputs of the last four hierarchies, because focal slice features from the first hierarchy often contain much cluttered information. All channel numbers of side outputs are then unified to 64 via the compression modules (CP) to facilitate computation. Let the compressed focal stack features be $(F_{fs}^{j}(i))_{i=2}^{5}$ ($j=1,2,...,12$), where $j$ means focal slice index, the compressed AiF features be $(F_{a}(i))_{i=2}^{5}$, and the compressed depth features be $(F_{d}(i))_{i=2}^{5}$. Features of the three modalities in the same hierarchy are simultaneously fed to a guided refinement and fusion module (GRFM) to refine the focal stack features $F_{fs}(i)$ and then aggregate the cross-modal features. Details of the module are described below.

\subsection{Guided Refinement and Fusion Module (GRFM)}
GRFM has two functions, \emph{i.e.}, refinement of the focal stack by AiF and depth information, and aggregation of multi-modal features. As shown in Fig.\:\ref{fig_blockdiagram}, for the first part, $F_{fs}(i)\in \mathbb{R}^{12 \times H \times W \times C}$ is split into dual streams by two $BConv$ for AiF-based and depth-based refinement, respectively. $BConv$ consists of convolution, BatchNorm and ReLU. Meanwhile, the extracted $F_{a}(i)$ and $F_{d}(i)\in \mathbb{R}^{1 \times H \times W \times C} $ are processed via channel attention (CA). Then, $F_{a}(i)$ and $F_{d}(i)$ are fed to two different refinement modules, each taking one branch of the focal stack features, respectively. The above process can be described as: 
\begin{align}
  F_{f}^{d}(i) = DRM(F_{d}(i), F_{fs}(i)), \label{DRM} \\
  F_{f}^{a}(i) = ARM(F_{a}(i), F_{fs}(i)), \label{ARM}
\end{align}
where $ARM$ and $DRM$ denote the two refinement modules, which will be described in the following paragraphs. It should be noted that, $F_{f}^{d}(i)$ and $F_{f}^{a}(i)\in \mathbb{R}^{1 \times H \times W \times C}$ are the results after guided refinement and slice fusion, and the slice fusion fuses the focal stack containing multiple slices across the batch dimension. Then, $F_{f}^{d}(i)$ and $F_{f}^{a}(i)$ are integrated by the cross-fusion operation (CF in Fig. \ref{fig_blockdiagram}) as below:
\begin{align}\label{equ:CF}
  F_{f}(i) = BConv[F_{f}^{d}(i) \oplus F_{f}^{a}(i), F_{f}^{d}(i) \otimes F_{f}^{a}(i)],
\end{align}
\noindent where $[\cdot,\cdot]$ denotes channel concatenation, $\oplus$ and $\otimes$ represent element-wise addition and multiplication, respectively. In order to make full use of the information from AiF and depth features, we finally perform a tri-modal aggregation operation, which is denoted as: 
\begin{align}\label{equ:Tri}
  F(i) = BConv[F_{f}(i), BConv[F_{a}(i), F_{d}(i)]].
\end{align}
The obtained outputs $F(i) (i \in \{2,3,4,5\})$ are fused across levels by $BConv$ in a UNet top-down manner as in Fig.\:\ref{fig_blockdiagram}. The ablation experiments in Sec.\:\ref{sec44} will validate the above design ideas of GRFM. 

\subsection{AiF-based Refinement Module (ARM)}
Details of ARM are shown in the orange box of Fig.\:\ref{fig_blockdiagram}. ARM first utilizes AiF features to weigh different slices of the focal stack. Then, the fused spatial attention masks are leveraged to filter noises on the slices, and finally different slices are integrated across the batch dimension. 

As shown in the orange box of Fig.\:\ref{fig_blockdiagram}, firstly, we design an alignment weighting module (AWM) to weigh for different slices by aligning them with the AiF feature tensors. The AiF image with abundant spatial structure information is more likely to focus on salient areas during the training process, like in the 2D SOD cases \cite{rgb2, rgb9, rgb10}. Therefore, a slice in the focal stack with higher alignment to the AiF features should contain more complete salient objects. Specifically, $F_{a}(i)$ and $F_{fs}(i)$ are processed separately via $BConv$ to obtain alignment-related features $W_{fs}(i)\in \mathbb{R}^{12 \times H \times W \times C}$ and $W_{a}(i)$. Moreover, we replicate $W_{a}(i)$ features by 12 times across the batch dimension to make it identical to the dimension of $W_{fs}(i)$. 
Inspired by Dice coefficient \cite{22}, we design a measure for the alignment degree:
\begin{align}\label{equ:BA}
  V = Softmax(Conv_{1}(\frac{GAP(W_{a}(i) \otimes W_{fs}(i))}{GAP(W_{a}(i) \oplus W_{fs}(i))})),
\end{align}
\noindent where $GAP$ is the global average pooling, and $Conv_{1}$ represents a shared $1 \times 1$ convolutional layer to compress channel numbers to 1. The obtained tensor $V \in \mathbb{R}^{12 \times 1 \times 1 \times 1}$ indicates the weights of different slices. The focal stack after weighting is obtained by multiplication as: $F_{fs}(i)^{1} = F_{fs}(i) \times V$, which pays more attention to the slices focused at the depth where  salient objects are located. Subsequently, the AiF features after replication, namely $F_{a}(i)^{1}$, and $F_{fs}(i)^{1}$ are used to generate the spatial attention masks including spatial and texture information to purify focal slices:
\begin{align}\label{equ:ARM_1}
  Mask = Sigmoid(Conv3(F_{fs}(i)^{1} \otimes F_{a}(i)^{1})),
\end{align}
\vspace{-0.6cm}
\begin{align}\label{equ:ARM_2}
  F_{fs}(i)^{2} = F_{fs}(i)^{1} \oplus (F_{fs}(i)^{1} \otimes Mask),
\end{align}
\noindent where $Conv3$ represents a series of $3 \times 3$ convolutions. Finally, a focal slice fusion operation (FF) is performed on all slice features in the focal stack, as shown in the red box of Fig.\:\ref{fig_blockdiagram}. To better integrate the cues from different slices, we apply both element-wise addition and maximization in FF:
\begin{align}\label{FF_1}
  F_{sum} = \sum_{j=1}^{12}(F_{fs}^{j}(i)^{2}), 
  F_{max} = Max(F_{fs}^{j}(i)^{2}),
\end{align}
\vspace{-0.5cm}
\begin{align}\label{FF_2}
F_{f}^{a}(i) = BConv[F_{sum}, F_{max}]. 
\end{align}
\noindent $F_{f}^{a}(i)$ is the refinement output of ARM. The ablation experiments in Sec.\:\ref{sec44} indicate that ARM can further improve the performance of SOD.

\begin{table*}[t!]
    \renewcommand{\arraystretch}{1.0}
    \caption{\small Quantitative results. The best result is highlighted in \textbf{bold}. ``N/T'' denotes unavailable results. $\uparrow$/$\downarrow$ means that a larger/smaller value is better.
    }\label{comparison}
   \centering
    \footnotesize
    \setlength{\tabcolsep}{2.1mm}
    \begin{tabular}{p{0.2mm}p{0.4mm}r||c|c|c|c|c|c|c|c|c|c|c|c||c}
    \hline
    & & & \multicolumn{4}{c|}{Traditional} &\multicolumn{9}{c}{Deep Learning}\\  
    \cline{4-16}
    
           \multicolumn{2}{p{0.4mm}}{Dataset}  & Metric  & \tabincell{c}{LFS\\\cite{fs_tra1}} & \tabincell{c}{WSC\\\cite{fs_tra2}} & \tabincell{c}{DILF\\\cite{fs_tra6}}  & \tabincell{c}{RDFD\\\cite{fs_tra5}} &  \tabincell{c}{MoLF\\\cite{fs7}} & \tabincell{c}{MAC\\\cite{fs6}} & \tabincell{c}{LFNet\\\cite{fs8}} & \tabincell{c}{ERNet\\\cite{fs9}} &  \tabincell{c}{PANet\\\cite{fs10}}& \tabincell{c}{TCFANet\\\cite{fs14}}&
           \tabincell{c}{SANet\\\cite{fs12}}& \tabincell{c}{DLGLRG\\\cite{fs11}}& Ours\\
    \hline
    \hline
          \multirow{4}{*}{\begin{sideways}\textit{DUTLF-FS}\end{sideways}} & \multirow{4}{*}{\begin{sideways}\cite{fs4}\end{sideways}} & $S_\alpha\uparrow$       &0.585&0.656&0.725&0.658&0.887&0.804&0.878&0.900&0.910&0.914&0.918&0.929&\textbf{0.931}\\
                                                                        && $F_{\beta}^{\textrm{max}}\uparrow$       &0.533&0.617&0.671&0.599&0.903&0.792&0.891&0.908&0.912&0.912&0.927&0.938&\textbf{0.941}\\
                                                                        && $E_{\phi}^{\textrm{max}}\uparrow$       &0.711&0.788&0.802&0.773&0.939&0.863&0.930&0.949&0.944&0.951&0.956&0.961&\textbf{0.965}\\
                                                                        && $M\downarrow$         &0.228&0.151&0.157&0.192&0.052&0.103&0.054&0.040&0.038&0.038&0.032&0.030&\textbf{0.026}\\
    \hline
        \multirow{4}{*}{\begin{sideways}\textit{HFUT-Lytro}\end{sideways}} & \multirow{4}{*}{\begin{sideways}\cite{fs3}\end{sideways}}& $S_\alpha\uparrow$
        &0.565&0.613&0.672&0.619&0.742&0.731&0.736&0.778&0.796&\textbf{0.816}&0.784&0.771&0.803\\
                                                                        && $F_{\beta}^{\textrm{max}}\uparrow$      &0.427&0.508&0.600&0.533&0.662&0.667&0.656&0.722&0.741&0.724&0.743&0.702&\textbf{0.756}\\
                                                                        && $E_{\phi}^{\textrm{max}}\uparrow$     &0.637&0.695&0.748&0.712&0.811&0.796&0.799&0.841&0.848&0.832&0.831&0.843&\textbf{0.852}\\
                                                                        && $M\downarrow$       &0.222&0.154&0.151&0.215&0.095&0.108&0.093&0.082&0.074&\textbf{0.068}&0.078&\textbf{0.068}&0.072\\

    \hline
        \multirow{4}{*}{\begin{sideways}\textit{LFSD}\end{sideways}}& \multirow{4}{*}{\begin{sideways}\cite{fs2}\end{sideways}} & $S_\alpha\uparrow$ &0.681&0.702&0.811&0.786&0.825&0.789&0.820&0.830&0.834&0.821&0.841&0.856&\textbf{0.857}\\
                                                                        && $F_{\beta}^{\textrm{max}}\uparrow$       &0.744&0.743&0.811&0.802&0.824&0.788&0.821&0.842&0.828&0.810&0.840&0.854&\textbf{0.859}\\
                                                                        && $E_{\phi}^{\textrm{max}}\uparrow$     &0.809&0.789&0.861&0.851&0.880&0.836&0.885&0.884&0.876&0.834&0.893&0.893&\textbf{0.906}\\
                                                                        && $M\downarrow$       &0.205&0.150&0.136&0.136&0.092&0.118&0.092&0.083&0.082&0.087&0.084&0.072&\textbf{0.065}\\
    \hline
        \multirow{4}{*}{\begin{sideways}\textit{Lytro Illum}\end{sideways}} & \multirow{4}{*}{\begin{sideways}\cite{fs6}\end{sideways}}  & $S_\alpha\uparrow$       &0.619&0.709&0.756&0.738&0.834&N/T&N/T&0.844&0.857&N/T&0.840&0.872&\textbf{0.889}\\
                                                                        && $F_{\beta}^{\textrm{max}}\uparrow$     &0.545&0.662&0.697&0.696&0.820&N/T&N/T&0.827&0.828&N/T&0.807&0.845&\textbf{0.875}\\
                                                                        && $E_{\phi}^{\textrm{max}}\uparrow$       &0.721&0.804&0.830&0.816&0.908&N/T&N/T&0.910&0.911&N/T&0.895&0.918&\textbf{0.932}\\
                                                                        && $M\downarrow$       &0.197&0.115&0.132&0.142&0.065&N/T&N/T&0.057&0.053&N/T&0.054&0.045&\textbf{0.037}\\
                                                                                                                                                                   
\hline
\end{tabular}
\vspace{-0.3cm}
\end{table*}

\subsection{Depth-based Refinement Module (DRM)}
Coordinate attention proposed in \cite{21} is a solid improvement upon the previous channel attention \cite{17} to preserve spatial attentive information. It factorizes the channel attention into two different directions, capturing long-range information in one direction and meanwhile preserving positional information in the other. 
Considering that coordinate attention is only designed to extract information from a single modality, we design DRM based on a modified coordinate attention to capture multi-modal long-range dependencies across the depth and focal stack. The intuition is that the depth map can provide useful positional clues, which can generate attention masks used for enhancing position representation of objects on different slices. 

Specifically, we first utilize two pairs of average pooling (\emph{i.e.}, ``Xpool'' and ``Ypool'' in DRM as shown in Fig.\:\ref{fig_blockdiagram}) to encode the focal stack and depth features along horizontal and vertical direction, respectively. Note the depth tensor is also replicated across the batch dimension (\emph{i.e.}, $F_{d}(i)^{1}$) to be aligned to $F_{fs}(i) \in \mathbb{R}^{12 \times H \times W \times C}$. Then, the obtained pairs of tensors (\emph{i.e.}, $X_{fs}(i)$ and $X_{d}(i) \in \mathbb{R}^{12 \times H \times 1 \times C}$, $Y_{fs}(i)$ and $Y_{d}(i) \in \mathbb{R}^{12 \times 1 \times W \times C}$) are cascaded along channel dimension, and fed to a $1 \times 1$ convolution to compress their channels to $C$. To better encode positional information, similar to \cite{21}, the two fused tensors are also cascaded and processed by $BConv1$ ($1 \times 1$ convolution with  BatchNorm and ReLU), as shown in Fig.\:\ref{fig_blockdiagram}. Finally, the cascaded features are split back into two tensors with the same dimension as the $X_{fs}(i)$ and $Y_{fs}(i)$, and then transformed to attention masks $X(i)$ and $Y(i)$ via $Conv1$ and $Sigmoid$. The focal slices are then refined and fused as follows:
\begin{align}\label{loss}
  F_{f}^{d}(i) = FF(F_{fs}(i) \otimes X(i) \otimes Y(i)),
\end{align}
\noindent where $FF$ denotes the same focal slice fusion operation (FF) indicated by Eq.\:\ref{FF_1} and Eq.\:\ref{FF_2} as aforementioned in ARM.

\subsection{Loss Function}
We adopt a combination of BCE (binary cross-entropy) loss, IOU (intersection-over-union) loss and 1-$E_{\phi}^{\textrm{max}}$ \cite{fs12} for training. The total loss is formulated as: \begin{equation}\label{equ:DRM}
\mathcal{L}_{total}=\sum_{i=2}^{5}\mathcal{L}(S_{i}, G) + \mathcal{L}(S_{AiF}, G) + \mathcal{L}(S_{dep}, G), 
\end{equation}
\noindent where $\mathcal{L}$ represents the combined loss, and $G$ is the ground truth. $S_{AiF}$, $S_{dep}$ and $S_{i}$ denote the outputs of the AiF branch, the depth branch and the decoder, respectively. Note that we have ignored the coarse prediction from the focal stack branch because individually fused focal slices without refinement contain much interference, leading to ineffective supervision ($\sim 1.3\%$ F-measure drop in average). During inference, $S_{2}$ is taken as the final predication. 

\section{Experiments And Results}
We evaluate GFRNet on four widely used light field datasets: DUTLF-FS \cite{fs4}, HFUT-Lytro \cite{fs3}, LFSD \cite{fs2} and Lytro Illum \cite{fs6}. For fair comparison, we follow the training protocol of \cite{fs11,fs9}, which selects 1,000 samples from DUTLF-FS and 100 samples from HFUT-Lytro for training. The remaining samples together with other datasets are for testing. Moreover, we adopt four common metrics to evaluate the performance of different models in a comprehensive way, including max F-measure ($F^{\textrm{max}}_{\beta}$) \cite{1}, S-measure ($S_{\alpha}$)\cite{12}, max E-measure ($E^{\textrm{max}}_{\phi}$) \cite{13} and Mean Absolute Error ($M$) \cite{14}.

\subsection{Implementation Details}
We implement GFRNet by Pytorch and all experiments are conducted on one RTX 2080Ti GPU. All input images are resized to 256 $\times$ 256. 
We adopt the Adam algorithm \cite{11} to train our model with batch size 1. The entire training process takes 50 epochs. The initial learning rate is set to 1e-5 and decayed by 10\% at the 40th epoch. The training data is augmented with flipping, random cropping and rotation.

\subsection{Comparisons to State-of-the-Arts}
For comprehensive comparison, we compare our method with 12 state-of-the-art (SOTA) light field methods, \emph{i.e.,} LFS\cite{fs_tra1}, WSC\cite{fs_tra2}, DILF\cite{fs_tra6}, RDFD\cite{fs_tra5}, MoLF\cite{fs7}, MAC\cite{fs6}, LFNet\cite{fs8}, ERNet\cite{fs9}, PANet\cite{fs10}, TCFANet\cite{fs14}, SANet\cite{fs12}, and DLGLRG\cite{fs11}. Several methods (\emph{e.g.,} DLGLRG\cite{fs11}, PANet\cite{fs10}) used 93 samples in LFSD for evaluation. We unify the sample number of LFSD to 100. Quantitative results in Table\:\ref{comparison} show that our method achieves the overall best results on almost all metrics. Note that our GFRNet performs particularly well on Lytro Illum, where many challenging samples are included, \emph{e.g.,} complex scenes, small objects, and weak object texture. 

Several visual comparisons between GFRNet and SOTAs are shown in Fig.\:\ref{fig_comparison}. It can be observed that GFRNet achieves better accuracy in challenging scenarios, \emph{e.g.,} cluttered background ($1^{st}$ row), low contrast ($2^{nd}$ row), multiple objects ($3^{rd}$ row), and interfering background ($4^{th}$ row).

\begin{figure}
\centerline{\includegraphics[width=1.0\columnwidth]{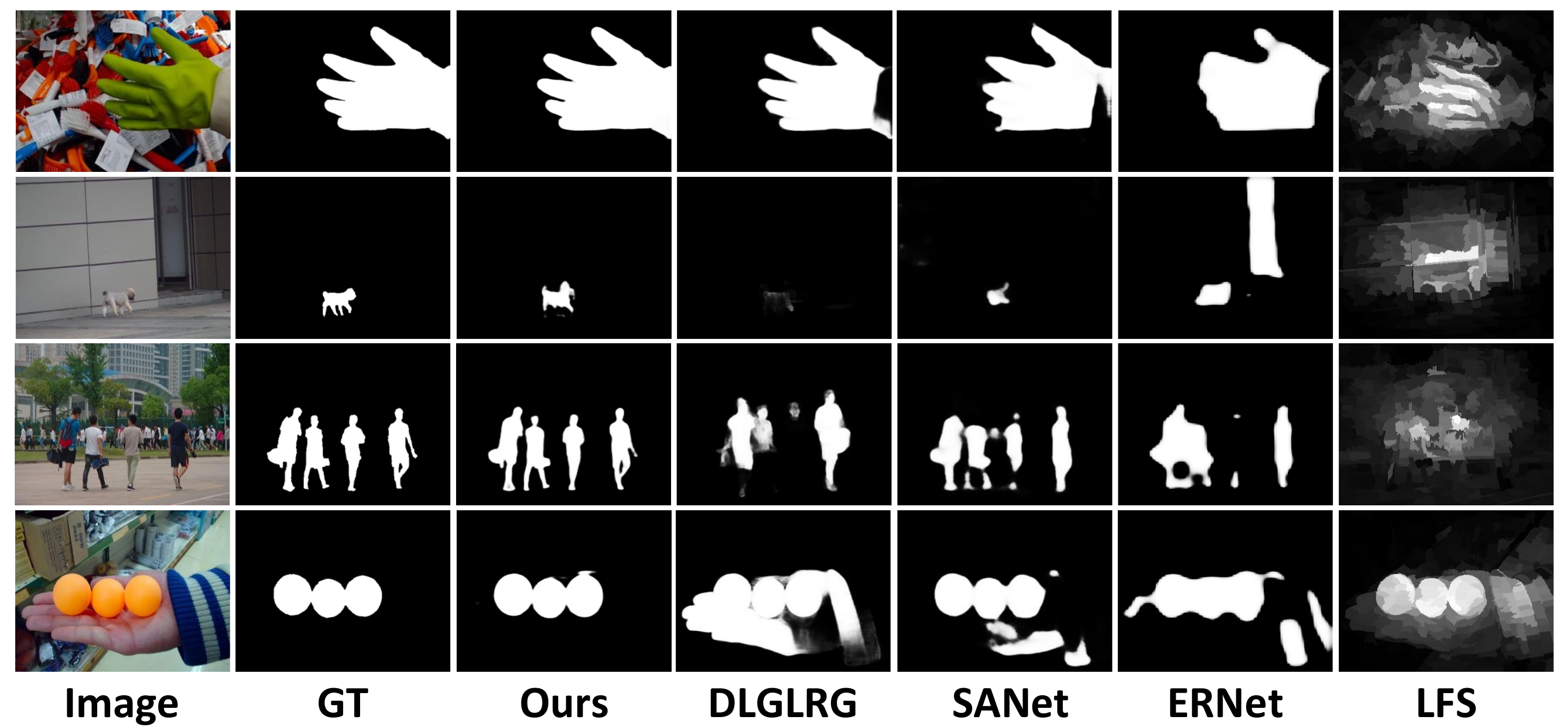}}\vspace{-0.3cm}
\caption{\small Visual comparison of our GFRNet and other SOTA models.}
\label{fig_comparison}
\end{figure}

\begin{table}[t!]
    \caption{\small Ablation study for ARM and DRM.}
    \label{tab:ablation_1}
    \renewcommand{\arraystretch}{1.0} 
    \centering
    \footnotesize
    \setlength{\tabcolsep}{2.8mm}
    
	 \begin{tabular}{c|cc|cc|cc}
		\hline
         \multirow{2}{*}{Variant} &
         \multicolumn{2}{c|}{Modes}
	   & \multicolumn{2}{c|}{DUTLF-FS~\cite{fs4}} 
	   & \multicolumn{2}{c}{Lytro Illum~\cite{fs6}}\\
	   \cline{2-7}
	   & AiF & Dep
	   & $F_{\beta}^{\textrm{max}}\uparrow$ & $S_\alpha\uparrow$
	   & $F_{\beta}^{\textrm{max}}\uparrow$ & $S_\alpha\uparrow$
		\\
	    \hline
	    M0  &\ding{54} &\ding{54}  & 0.919 & 0.916 & 0.842 & 0.868 \\
	    M1  & $A$ &\ding{54} & 0.930 & 0.923 & 0.860 & 0.878 \\
	    M2  &\ding{54} & $D$ & 0.923 & 0.919 & 0.853 & 0.873 \\
        M3  & $A$ & $A$ & 0.937 & 0.927 & 0.866 & 0.881 \\
        M4  & $D$ & $D$ & 0.928 & 0.921 & 0.862 & 0.878 \\
        M5  & $D$ & $A$ & 0.926 & 0.922 & 0.859 & 0.875 \\
        \rowcolor{gray!40} Full  & $A$ & $D$ & \textbf{0.941} & \textbf{0.931} & \textbf{0.875} & \textbf{0.889} \\
        \hline
    \end{tabular}
 \vspace{-0.3cm}
\end{table}

\subsection{Ablation Study}\label{sec44}
To validate our GFRNet, we conduct ablation experiments on two main datasets, \emph{i.e.}, DUTLF-FS and Lytro-Illum.

\textbf{Effectiveness of ARM and DRM.} To demonstrate the rationality of adopting modality-specific focal stack refinement strategies for AiF and depth branches, we set up experiments on six model variants. As shown in Table\:\ref{tab:ablation_1}, notations ``AiF'' and ``Dep'' represent AiF-based and depth-based refinement paths, respectively. $A$ and $D$ denote ARM and DRM at specific positions, and \ding{54} denotes the setting which uses the focal slice fusing operation (FF in Fig.\:\ref{fig_blockdiagram}) to replace them. It can be seen that both ARM ($A$) and DRM ($D$) can help the model improve accuracy. Furthermore, AiF-based and depth-based refinement strategies corresponding to ARM and DRM, respectively, can achieve the best results (namely the ``Full'' model). Moreover, ``M5'' illustrates that the exchanged use of ARM and DRM will degrade the accuracy. For ``M0'' and ``Full'' settings, we have visualized the focal stack features $F_{f}(4)$ shown in GRFM of Fig.\:\ref{fig_blockdiagram} for comparison and list three sets of examples, which are shown in Fig.\:\ref{fig_visualization}. One can see that focal stack features with refinement contain much fewer impurities and focus more on salient objects.

\begin{figure}
\centerline{\includegraphics[width=1.0\columnwidth]{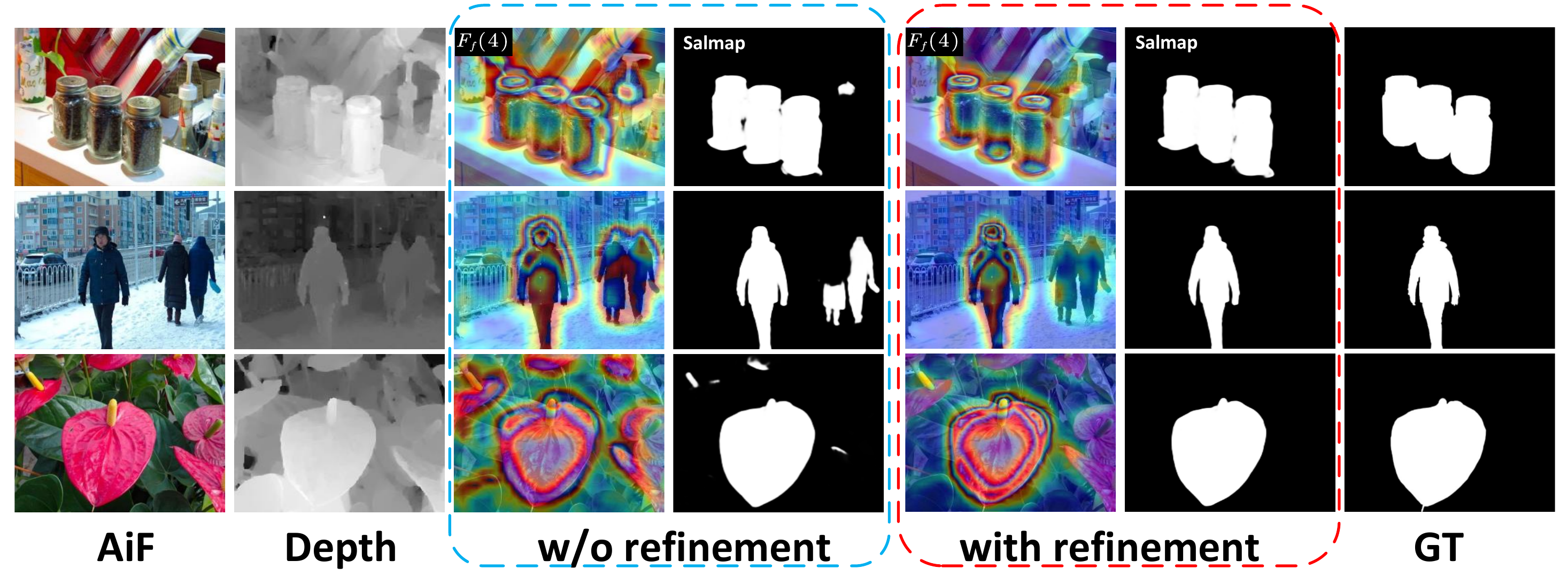}}\vspace{-0.2cm}
\caption{\small Visualized feature maps (``$F_{f}(4)$'' in GRFM of Fig.\:\ref{fig_blockdiagram}) and saliency maps, under the settings ``M0'' (without refinement) and ``Full'' (with refinement). GT means ground truth.}
\label{fig_visualization}
\end{figure}

\begin{table}[t!]
    \caption{\small Ablation study for multi-modal refinement.}
    \label{tab:ablation_2}
    \renewcommand{\arraystretch}{1.0} 
    \centering
    \footnotesize
    \setlength{\tabcolsep}{2.2mm}
    
	 \begin{tabular}{c|ccc|cc|cc}
		\hline
         \multirow{2}{*}{Variant} &
         \multicolumn{3}{c|}{Modalities}
	   & \multicolumn{2}{c|}{DUTLF-FS~\cite{fs4}} 
	   & \multicolumn{2}{c}{Lytro Illum~\cite{fs6}}\\
	   \cline{2-8}
	   & FS & AiF & Dep
	   & $F_{\beta}^{\textrm{max}}\uparrow$ & $S_\alpha\uparrow$
	   & $F_{\beta}^{\textrm{max}}\uparrow$& $S_\alpha\uparrow$
		\\
	    \hline
	    V0  & \ding{52} & & & 0.877 & 0.887 & 0.801 & 0.838 \\
	    V1  & \ding{52} & \ding{52} & & 0.933 & 0.920 & 0.857 & 0.876 \\
            V2  & \ding{52} & & \ding{52} & 0.915 & 0.913 & 0.840 & 0.861 \\
        \rowcolor{gray!40} Full  & \ding{52} & \ding{52} & \ding{52} & \textbf{0.941} & \textbf{0.931} & \textbf{0.875} & \textbf{0.889} \\
        \hline
    \end{tabular}
\end{table}

\textbf{Effectiveness of multi-modal refinement.} We conduct three variants to validate the solidity of using AiF and depth information to refine focal stacks and enhance saliency detection. As shown in Table.\:\ref{tab:ablation_2}, ``FS'', ``AiF'' and ``Dep'' represent the three modal branches, respectively. Setting ``V0'' just remains the focal stack branch, and uses the focal slice fusing operation (FF in Fig.\:\ref{fig_blockdiagram}) to replace the refinement modules. ``V1'' complements ``V0'' with the AiF branch including ARM, and aggregates focal stack and AiF features by a convolutional layer instead of the tri-modal aggregation operation in GRFM of Fig.\:\ref{fig_blockdiagram}. ``V2'' adopts the same construction as ``V1'', but adds the depth branch with DRM instead of the AiF branch. Comparing ``V1'' and ``V2'' with ``V0'', respectively, one can find that both AiF and depth information can help focal stacks to boost the performance of SOD. The full design of GFRNet also illustrates that the application of AiF and depth together can achieve the best results.

\textbf{Other design details of GRFM.} We conduct different variants to validate GRFM's design details. Setting ``P0'' replaces the cross-fusion operation (CF) with a simple concatenation operation. ``P1'' denotes the variant where we replace the focal slice fusion operation (FF) with a convolutional operation after cascading focal slices along channel axis. ``P2'' uses a convolutional layer after direct concatenation to replace the tri-modal aggregation operation, while the latter utilizes two convolutional layers to achieve progressive fusion. The results are summarized in Table.\:\ref{tab:ablation_3}. It can be observed that these variants achieve worse results compared with the full model, validating the effectiveness of these detailed designs.

\begin{table}[t!]
    \renewcommand{\arraystretch}{1.0}
    \caption{\small Ablation for detailed designs of GRFM.}
    \label{tab:ablation_3}
    \renewcommand{\arraystretch}{1.0} 
    \centering
    \footnotesize
    \setlength{\tabcolsep}{2.5mm}
	 \begin{tabular}{c|ccc|ccc}
		\hline
         \multirow{2}{*}{Variant} &
		 \multicolumn{3}{c|}{DUTLF-FS~\cite{fs4}} 
	   & \multicolumn{3}{c}{Lytro Illum~\cite{fs6}}\\
	   \cline{2-7} 
	   & $M\downarrow$ & $F_{\beta}^{\textrm{max}}\uparrow$ & $S_\alpha\uparrow$
	   & $M\downarrow$ & $F_{\beta}^{\textrm{max}}\uparrow$ & $S_\alpha\uparrow$
		\\
	    \hline
	    P0 & 0.027 & 0.937 & 0.929 & 0.039 & 0.874 & 0.887 \\
        P1 & 0.027 & 0.940 & 0.926 & 0.041 & 0.866 & 0.880 \\
        P2 & 0.028 & 0.936 & 0.928 & 0.039 & \textbf{0.875} & 0.887 \\
        \rowcolor{gray!40} Full & \textbf{0.026} & \textbf{0.941} & \textbf{0.931} & \textbf{0.037} & \textbf{0.875} & \textbf{0.889} \\
        \hline
    \end{tabular}
    \vspace{-0.3cm}
\end{table}

\section{Conclusion}
We propose a novel guided focal stack refinement network called GFRNet, by refining the focal stack to improve light field salient object detection. We design a new guided refinement and fusion module (GRFM) plugged into the network to refine focal stack features and aggregate multi-modal features. Further in GRFM, we elaborately design an AiF-based refinement module (ARM) and a depth-based refinement module (DRM) to leverage their complementary properties in a guided manner. The effectiveness of these key designs has been validated by comprehensive ablation analyses.
\\ \hspace*{\fill} \\

{
\bibliographystyle{IEEEbib}
\bibliography{GFRNet}
}

\end{document}